%% file: example.tex
\documentclass[letterpaper, 10 pt, conference]{ieeeconf}  % Comment this line out if you need a4paper

\IEEEoverridecommandlockouts                              % This command is only needed if 
\usepackage[utf8]{inputenc} % allow utf-8 input
\usepackage[T1]{fontenc}    % use 8-bit T1 fonts
\usepackage{hyperref}       % hyperlinks
\usepackage{url}            % simple URL typesetting
\usepackage{booktabs}       % professional-quality tables
\usepackage{amsfonts}       % blackboard math symbols
\usepackage{nicefrac}       % compact symbols for 1/2, etc.
\usepackage{microtype}      % microtypography

\usepackage{comment}
\usepackage{paralist}
\usepackage{graphicx}
\usepackage{amsmath}
\usepackage{cleveref}
\usepackage[english]{babel}
\usepackage{multirow}
\usepackage[skip=2pt]{caption} % example skip set to 2pt
\usepackage{subcaption}
\usepackage{sidecap}
\usepackage{floatrow}
\usepackage[affil-it]{authblk}
\usepackage{bbm}
\title{\LARGE \bf
Image-Based Conditioning for Action Policy Smoothness in Autonomous Miniature Car Racing with Reinforcement Learning
}

\author[1]{Bo-Jiun Hsu}
\author[1,2]{Hoang-Giang Cao}
\author[1]{I Lee}
\author[1]{Chih-Yu Kao}
\author[1]{Jin-Bo Huang}
\author[1,2]{I-Chen Wu$^\dagger$}
\affil[1]{Department of Computer Science, National Yang Ming Chiao Tung University, Taiwan}
\affil[2]{Research Center for IT Innovation, Academia Sinica, Taiwan}

\begin{document}

\maketitle

\def\thefootnote{$\dagger$}\footnotetext{Correspondence.}\def\thefootnote{\arabic{footnote}}

\thispagestyle{empty}
\pagestyle{empty}

%keywords
%deep reinforcement learning, autonomous driving, miniature car racing, smooth control, sim to real

%%%%%%%%%%%%%%%%%%%%%%%%%%%%%%%%%%%%%%%%%%%%%%%%%%%%%%%%%%%%%%%%%%%%%%%%%%%%%%%%
\begin{abstract}
In recent years, deep reinforcement learning has achieved significant results in low-level controlling tasks.
However, the problem of control smoothness has less attention.
In autonomous driving, unstable control is inevitable since the vehicle might suddenly change its actions.
This problem will lower the controlling system's efficiency, induces excessive mechanical wear, and causes uncontrollable, dangerous behavior to the vehicle. 
In this paper, we apply the Conditioning for Action Policy Smoothness (CAPS) with image-based input to smooth the control of an autonomous miniature car racing.
Applying CAPS and sim-to-real transfer methods helps to stabilize the control at a higher speed. 
Especially, the agent with CAPS and CycleGAN reduces 21.80\% of the average finishing lap time. 
Moreover, we also conduct extensive experiments to analyze the impact of CAPS components.

\end{abstract}

%%%%%%%%%%%%%%%%%%%%%%%%%%%%%%%%%%%%%%%%%%%%%%%%%%%%%%%%%%%%%%%%%%%%%%%%%%%%%%%%
\section{INTRODUCTION}

In recent years, deep reinforcement learning (DRL) has archived many milestones.
AlphaZero, a computer program based on reinforcement learning, achieved superhuman performance in chess, shogi, and Go through self-play\cite{Silver2018rl}.
For the real-time strategy (RTS) games, deep reinforcement learning also reached the human level performance. 
For example, in 2018, DeepMind proposed AlphaStar, and defeated professional players for the first time\cite{Vinyals2019Grandmaster};
OpenAI Five is the first AI to beat the world champions in an e-sports game \cite{openai2019dota}.
DRL is also applied to many real-world applications.
Shixiang Gu et al. applied DRL for robotic manipulation with asynchronous off-policy updates \cite{Gu2017off}.
% The work in \cite{PI2020104222} uses low-level control through reinforcement learning directly to motors output.
For autonomous driving, Amazon Web Services (AWS) provided DeepRacer, an autonomous racing experimentation platform for sim-to-real reinforcement learning\cite{Bharathan2019DeepRacer}.
However, training DRL usually requires a large amount of data for training.
Collecting data in the real world is extremely costly and time-consuming.
In addition, to acquire an optimal policy or near-optimal policy, the agent needs to perform random actions to explore, which can be unstable and dangerous during real-world data collection for a task like autonomous driving.
Sim-to-real provides an efficient approach for training DRL.
We often train agents in the simulation and migrate them to the real world.

However, the intrinsic gap between simulation and the real world often causes the agent trained in simulation to perform poorly in the real-world \cite{hofer2020perspectives}.
Prior works mainly focus on domain randomization and domain adaptation methods.
The most widely used sim-to-real transfer method is domain randomization, but domain randomization requires careful task-specific hyper-parameter tuning, and the cost grows exponentially with the number of parameters \cite{zhao2020sim,hofer2020perspectives}.
The work in \cite{chu2020sim} proposed a student-teacher method for sim-to-real transfer in DeepRacer.
The student model after training can perform better than the teacher model and other randomization methods.
Recent works turned to learning-based domain adaptation.
RL-CycleGAN combined Generative Adversarial Network (GAN) with a task-specific RL constraint in a robotics grasping task, but collecting task specific data can be expensive \cite{kan2020rlcyclegan}. 
RetinaGAN improved GAN via object detection consistency auxiliary loss\cite{ho2020retinagan}. However, in autonomous driving, the observation has no specific object for tracking to follow the perception consistency loss as proposed.
% In this paper, we will implement CycleGAN\cite{CycleGAN2017}, a domain adaptation method, to compare its effectiveness with other domain randomization methods.

Another critical problem in autonomous driving is the jerky behavior, where the vehicle moves unstable or shaky.
Jerky control causes many serious problems, such as uncontrollable moving and power-consuming, which may reduce the service life of the autonomous robot.
Siddharth Mysore et al. proposed Conditioning for Action Policy Smoothness (CAPS) for solving jerky actions by adding regularization terms\cite{siddharth2021caps}.
CAPS was originally applied to smooth the control of quadrotor drones.
There is less research about solving the jerky control problem in autonomous driving.

In this paper, we extend the idea of Conditioning for Action Policy Smoothness (CAPS)\cite{siddharth2021caps} to image-based CAPS and applying to solve jerky control in autonomous miniature car racings.
In the experiment, we analyze the impact of CAPS components.
We also compare the effectiveness of sim-to-real transfer approaches in autonomous car racings.

\textbf{Contributions.} The main contributions of this paper can be summarized as follow:
\begin{inparaenum}[1)]
    \item We extend CAPS\cite{siddharth2021caps} to image-based CAPS, and apply to solve jerky control in autonomous car racing.
    % \item  We compare the effectiveness between domain adaptation (CycleGAN) and domain randomization in solving  the sim-to-real problem in autonomous car racing.
    \item In the experiments, we show that CAPS helps stabilize the car when moving at a higher speed.
    Especially, the method that combines CycleGAN and CAPS outperforms other methods and reduces the average finish lap time, while also improving the completion rate.
    \item We also conducted extensive experiments to study the impact of CAPS components.
\end{inparaenum}

%%%%%%%%%OUR APPROACH%%%%%%%%%%%%%%%%%%%%%%
\section{CAPS: Conditioning for Action Policy Smoothness}
\label{sec:caps}
%review CAPS
Siddharth Mysore et al. introduced Conditioning for Action Policy Smoothness (CAPS)\cite{siddharth2021caps} and got significant improvements in controller smoothness and power consumption on a quadrotors drone.
To condition policies for smooth control, the authors proposed two regularization terms:
\begin{inparaenum}[1)]
    \item Temporal Smoothness term.
    \item Spatial Smoothness term.
\end{inparaenum}

%==== NOTATION=====

The policy $\pi$ is a mapping function of states $s$ to actions $a = \pi(s)$.
The objective function of CAPS, $J_{\pi}^{CAPS} $, contains three components: objective function of a model-free DRL $J_{\pi}$ (we used Soft Actor Critic in this work); Temporal Smoothness regularization term $L_{T}$; and Spatial Smoothness regularization term $L_{S}$.
The regularization weights $\lambda_{T}$ and $\lambda_{S}$ are used to balance the impact of two regularization terms $L_{T}$ and $L_{S}$, respectively.

\begin{equation}
J_{\pi}^{CAPS} = J_{\pi} - \lambda_{T}L_{T} - \lambda_{S}L_{S}
\label{equation:objective_functino_caps}
\end{equation}
\begin{equation}
L_{T} = D_{T}(\pi(s_{t}),\pi(s_{t+1}))
\end{equation}
\begin{equation}
L_{S} = D_{S}(\pi(s_{t}),\pi({s}_{t}')) \quad where \quad s_t'\sim \Phi(s_{t})
\end{equation}

$D$ is the Euclidean distance, i.e $D(a,b)= ||a-b||_{2}$.
%===== END NOTATION

\begin{figure}[t!]
\centering
    \begin{subfigure}[t]{0.45\columnwidth}
        \includegraphics[width=\columnwidth]{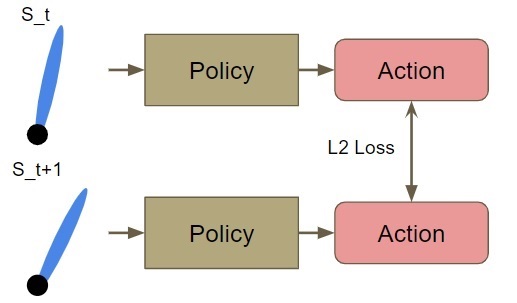}
    \caption{Temporal Smoothness.}
    \label{fig:temporal_smoothness}
    \end{subfigure}
    \begin{subfigure}[t]{0.45\columnwidth}
        \includegraphics[width=\columnwidth]{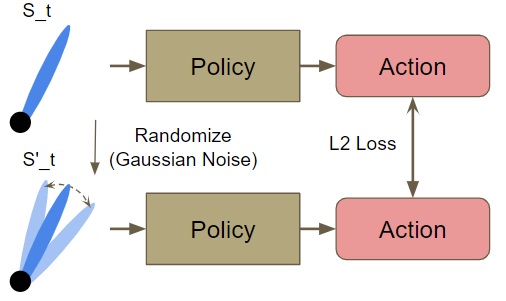}
        \caption{Spatial Smoothness.}
        \label{fig:spartial_smoothness}
    \end{subfigure}
\caption{Condition policies for smooth behavior in CAPS.}
\label{fig:caps_smoohness}
\end{figure}

The Temporal Smoothness term $L_{T}$ penalize the $J_{\pi}^{CAPS}$ when the action of the next states $s_{t+1}$ are significantly different from the actions of the current states $s_{t}$.
The Spatial Smoothness term $L_{S}$ ensures that the policy takes the similar actions on the similar states $S_t'$, which are drawn from a distribution $\phi$ around states $s$.
Originally, CAPS is applied to smooth the control of a quadrotor drone with the input are the inertial measurement unit (IMU) and the electronic speed controller (ESC) sensors.
Therefore, to generate similar states $S_t'$, the authors sample from normal distribution, \(\Phi(s) = N(s,\sigma)\) with standard deviation \(\sigma\), 
around $s$.
\autoref{fig:caps_smoohness} illustrates the Temporal Smoothness and Spatial Smoothness in CAPS.

In this paper, we extend the idea of CAPS to smooth the control of a miniature car racing with the image-based input.

\section{OUR APPROACH}
\subsection{Image-based CAPS for Autonomous Car Racing}
In this work, we extend the idea of using CAPS with image-based input and then apply it to smooth the control of an autonomous miniature car racing.
CAPS was originally applied to smooth the control of a drone with the internal states of the rotors. Then, to generate the similar state $S_t'$, the authors used Gaussian Noise to draw $S_t'$ from a normal distribution around state $S_t$ as describe in \autoref{sec:caps}.
Our approach use image as the input, therefore, to generate the similar state $S_t'$ in the Spatial Smoothness in CAPS, we implement 6 domain randomization methods:
\begin{enumerate}
    \item Random Brightness: adjust the brightness of the image.
    \item Random Contrast: adjust the degree to which light and dark colors in the image differ.
    \item Random Rotation: rotate the image with a random angle.
    \item Salt and Pepper: add salt and pepper noise to the image.
    \item Gaussian Blur: blur an image by a Gaussian function.
    \item Random Cut-off: random cut-off the image by overlaying a black rectangle at a random position with random size on the top of the image.
\end{enumerate}

\begin{figure}[!htb]
    \centering
        \includegraphics[width=0.9\columnwidth]{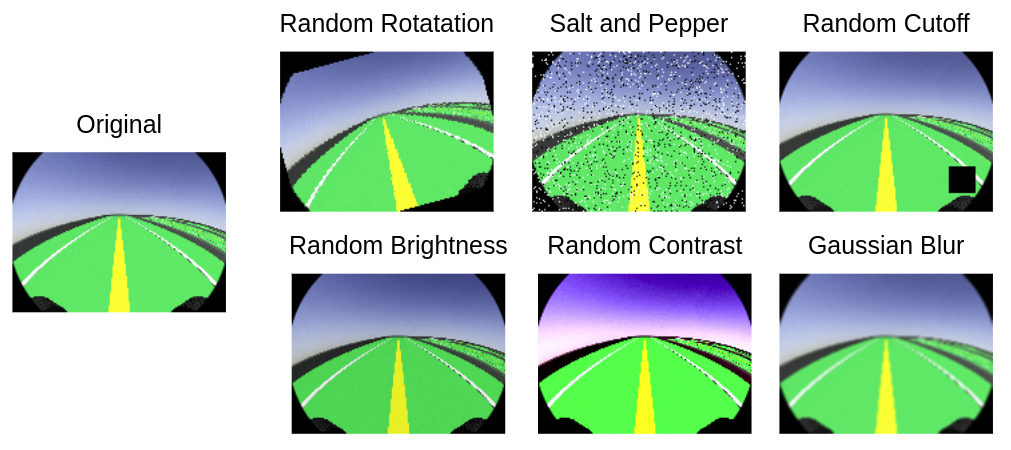}
    \caption{Domain randomization methods.}
    \label{fig:domain_randomization_result_2}
\end{figure}

\autoref{fig:domain_randomization_result_2} demonstrates our domain randomization methods that we used to generate the similar states $S_t'$ from states $s$.

We also study the impact between Temporal Smoothness and Spatial Smoothness by adjusting the regularization weights ${ \lambda{_T}}$ and \textbf{ ${\lambda{_S}}$} in the \autoref{subsec:Temporal_Smoothness_Sensitivity}

\subsection{Sim-to-real transfer for Autonomous Car Racing}
Sim-to-real transfer plays an important role in the sim-to-real task.
In sim-to-real tasks, we trained in the policy in the simulation and then directly transfer in the real world environment.
There are two main approaches for sim-to-real transfer: domain randomization and domain adaptation.
Domain randomization tries to enlarge the data distribution to cover the data as much as possible;
while domain adaptation tries to translate the data from this domain to another domain, which is simulation and real in this paper.

To analize the effectiveness of different sim-to-real transfer approaches in autonomous miniature car racings, we implement CycleGAN, a domain adaptation method, and three different domain randomization methods:
\begin{inparaenum}[1)]
    \item Salt and Pepper Noise \cite{bengio2011distributionexamples}.
    \item Random Reflection.
    \item Random HSV Shift.
\end{inparaenum}

%%%%%%%%%%%%%%%%%%%%%%%%%%%%%%%%%%%%%%%%%%%%%%%%%%%%%%%%%%%%%%%%%%%%%%%%%%%%%%%%
\section{EXPERIMENT SETUP}

\subsection{Environments Setup}

In the real world, our track had two straight acceleration zones, two square corners, three hairpin corners, and one s-curve.
If the car crosses the trapezoidal wall, it fails.

In the simulation, there is no reflections, noises, or car shake on the simulator track, while the real track has distinct differences because of the dynamic sunlight changing.

\begin{figure}[t!]
\centering
    \begin{subfigure}[t]{0.49\columnwidth}
        \includegraphics[width=\columnwidth]{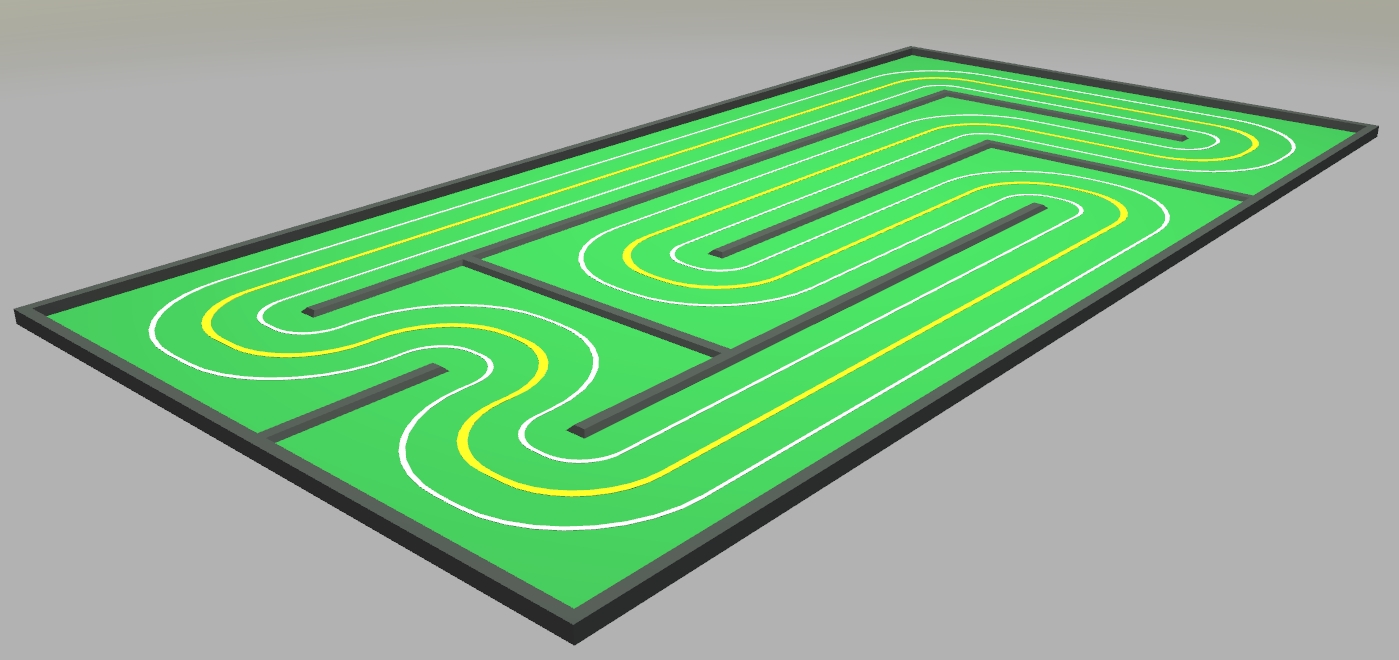}
    \caption{Simulation track.}
    \label{fig:sim_track}
    \end{subfigure}
    \begin{subfigure}[t]{0.49\columnwidth}
        \includegraphics[width=\columnwidth]{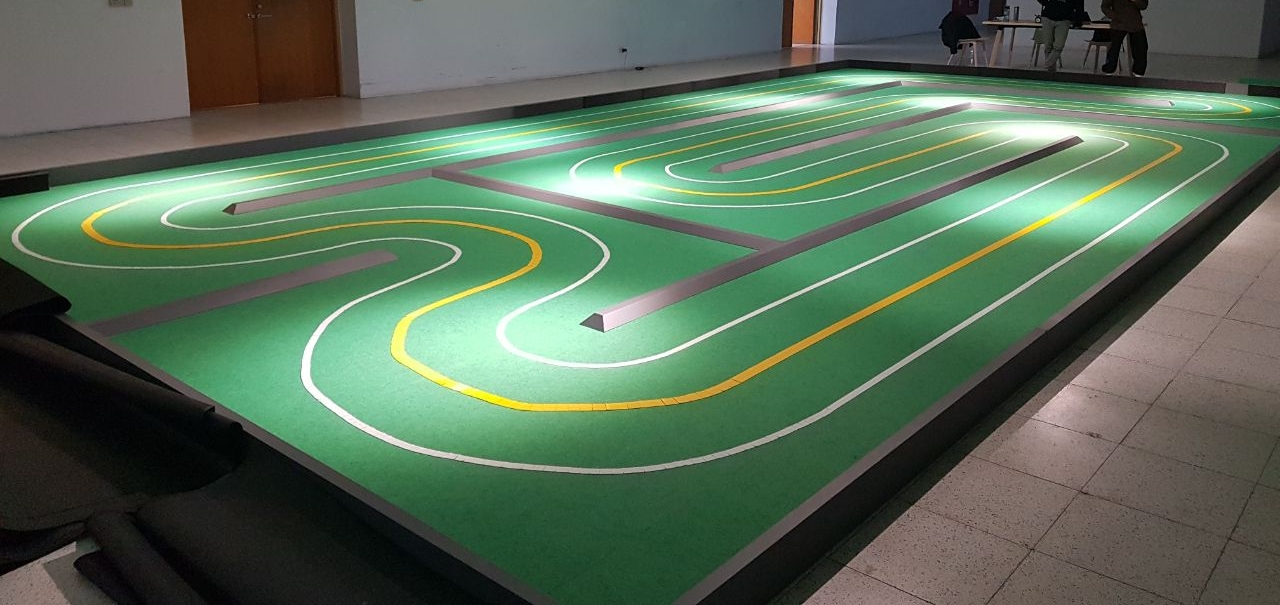}
        \caption{Real track.}
        \label{fig:real_track}
    \end{subfigure}
\caption{Simulation track(left) and real track(right).}
\label{fig:tracks}
\end{figure}

\begin{figure}[t!]
\centering
    \begin{subfigure}[t]{0.4\columnwidth}
        \includegraphics[width=\columnwidth]{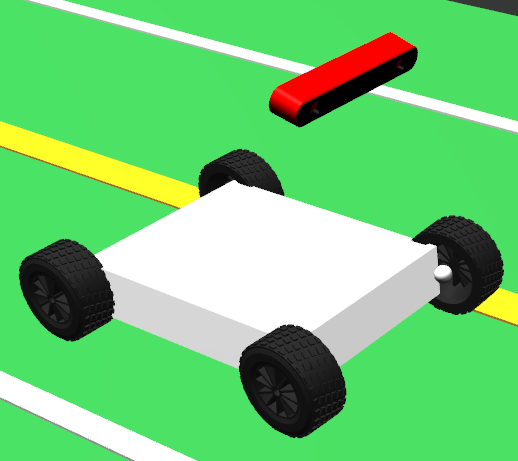}
    \label{fig:sim_car}
    \end{subfigure}
    \begin{subfigure}[t]{0.4\columnwidth}
        \includegraphics[width=\columnwidth]{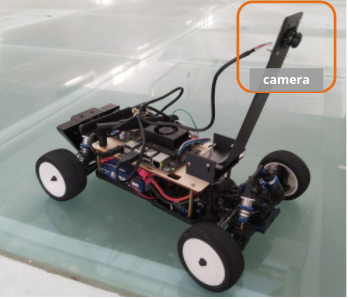}
        \label{fig:real_car}
    \end{subfigure}
\caption{Simulation car(left) and real car(right).}
\label{fig:cars}
\end{figure}

\autoref{fig:tracks} are our simulation track and real track.
The autonomous miniature car is set up with a camera placed on the top of the car. See \autoref{fig:cars}.
The camera captures RGB images with a resolution of 120x160 at 30fps.
The computing device used in the real car is NVIDIA Jetson Xavier NX.

\subsection{Network Structure and Training}
We implement a version of Ape-X\cite{Dan2018apex} with Soft-Actor-Critic (SAC)\cite{Haarnoja2018SAC} with the following configurations: number of workers is 3; N-step is 4; gamma is 0.98; initial alpha is 0.3; batch size is 512; learning rate is 0.0003; global buffer size is 45000; local buffer size is 2000.

\begin{figure}[!htb]
    \centering
        \includegraphics[width=0.95\columnwidth]{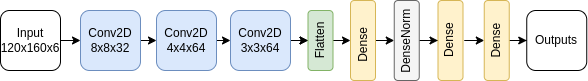}
    \caption{Network structure.}
    \label{fig:network_structure}
\end{figure}

Our input is the images that are captured from a camera placed on the top of the miniature car racing, \autoref{fig:cars}.
The observation is a RGB image with the resolution 120x160, and the color of each pixel ranges from 0 to 255.
We stack the current observation with the previous observation as the input. Therefore the input size is 120x160x6.
The output are the continuous values of steering angle and speed with a range in [-1, 1]. 
\autoref{fig:network_structure} illustrate our network structure used in this paper.

\subsection{Baselines}
We implement a version of Deep Neural Network Ape-X\cite{Dan2018apex} with Soft-Actor-Critic (SAC)\cite{Haarnoja2018SAC} without sim-to-real transfer method as the baseline.
We then compare the effectiveness of sim-to-real methods and CAPS for the smoothness control. The different models are as follows:
\begin{enumerate}
    \item Baseline with CAPS and no CAPS.
    \item Domain randomization with CAPS and without CAPS.
     \item CycleGAN with and without CAPS.
\end{enumerate}

\subsection{Metrics}
To evaluate the effectiveness of the smoothness method (CAPS), we use two following metrics:
\begin{enumerate}
    \item Mean of selected steering angle: measure the angle difference when the agent decide to change the moving direction.
    \item Smoothness value: proposed by \cite{siddharth2021caps}, a method based on the Fast Fourier Transform frequency spectrum as defined in \autoref{equation:smoothness_function} .
    The lower the value, the higher the low-frequency control and smoother the action.
\end{enumerate}

\begin{equation}
\label{equation:smoothness_function}
    S_{m} = \frac{2}{nf_{s}}\sum_{i=1}^{n}M_{i}f_{i}
\end{equation}
where $M_i$ is the amplitude of the frequency component $f_i$, and the $f_s$ is the sampling rate. We set $f_s=30$ in this work.

To evaluate the performance of the car racing, we use two metrics:
\begin{enumerate}
    \item Average finishing lap time: the average time to finish a run (in second).
    \item Completion rate: the \% number of completion runs per all runs.
\end{enumerate}

\section{EXPERIMENT RESULTS}
In the experiment, we first verify the performance of image-based CAPS in \ref{sec:smoothness_with_caps}. We then combine CAPS with different sim-to-real methods to compare the effectiveness of sim-to-real transfer for autonomous miniature car racing (\ref{subsec:caps_cyclegan}). Finally, we conduct extensive experiments to analyze the impact  of CAPS components (\ref{subsec:impact_CAPS}, \ref{subsec:Temporal_Smoothness_Sensitivity}, \ref{subsec:ablation_study}).

\subsection{Smoothness with CAPS}

\label{sec:smoothness_with_caps}
\begin{table}[]
\begin{tabular}{|l|l|l|}
\hline
& Without CAPS & With CAPS      \\ \hline
Steering angle            & 93.47        & \textbf{20.39} \\ \hline
Speed & 78.35        & \textbf{13.72} \\ \hline
\end{tabular}
\caption{Compare the smoothness value of steering angle and speed with CAPS and without CAPS.}
\label{tab:caps_vs_no_caps}
\end{table}

In this experiment, we study the smoothness ability of CAPS.
We train the agent in the simulation and test it in the real environment. The sim-to-real method is CycleGAN.

\begin{figure}[t!]
\centering
    \begin{subfigure}[t]{0.45\columnwidth}
        \includegraphics[width=\columnwidth]{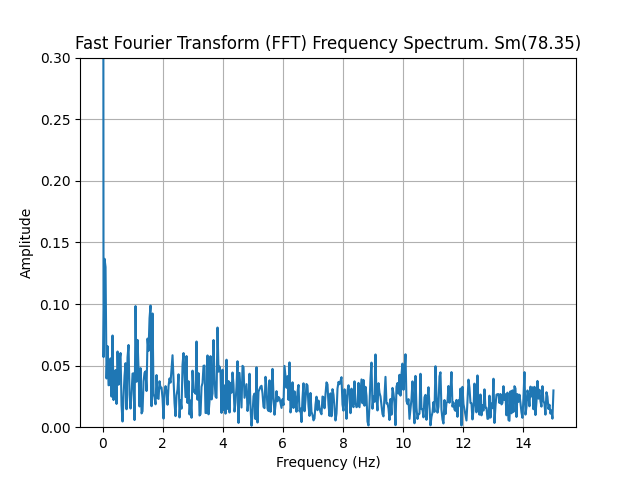}
        %\caption{Smoothness of speed without CAPS.}
    \label{fig:speed_smooth_no_caps}
    \end{subfigure}
    \begin{subfigure}[t]{0.45\columnwidth}
        \includegraphics[width=\columnwidth]{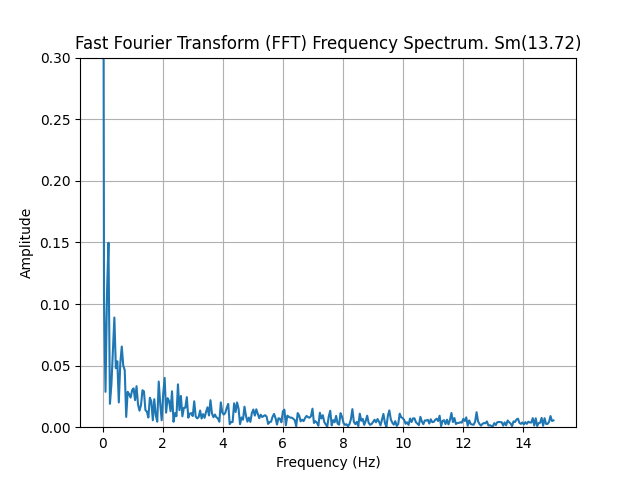}
        %\caption{Smoothness of speed with CAPS.}
        \label{fig:speed_smooth_caps}
    \end{subfigure}
\caption{Smoothness value of speed without CAPS (left) and with CAPS (right).}
\label{fig:compare_speed_smooth_caps}
\end{figure}

\begin{figure}[t!]
\centering
    \begin{subfigure}[t]{0.45\columnwidth}
        \includegraphics[width=\columnwidth]{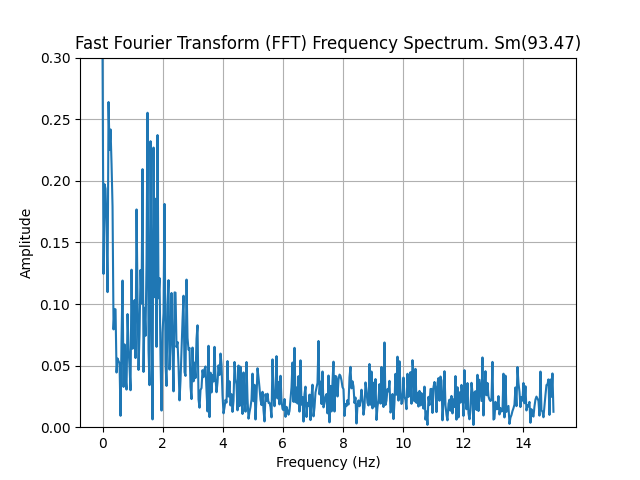}
        %\caption{Smoothness of steering angle without CAPS.}
    \label{fig:steering_smooth_no_caps}
    \end{subfigure}
    \begin{subfigure}[t]{0.45\columnwidth}
        \includegraphics[width=\columnwidth]{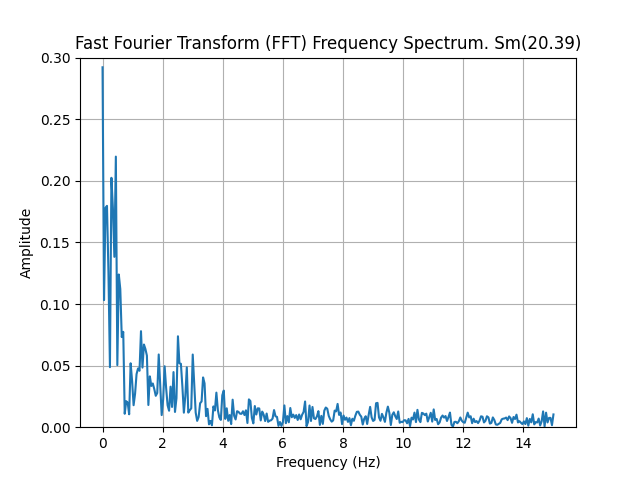}
        %\caption{Smoothness of steering angle with CAPS.}
        \label{fig:steering_smooth_caps}
    \end{subfigure}
\caption{Smoothness value of steering angle without CAPS (left) and with CAPS (right).}
\label{fig:compare_steering_smooth_caps}
\end{figure}

\begin{figure}[t!]
\centering
    \begin{subfigure}[t]{0.95\columnwidth}
        \includegraphics[width=\columnwidth]{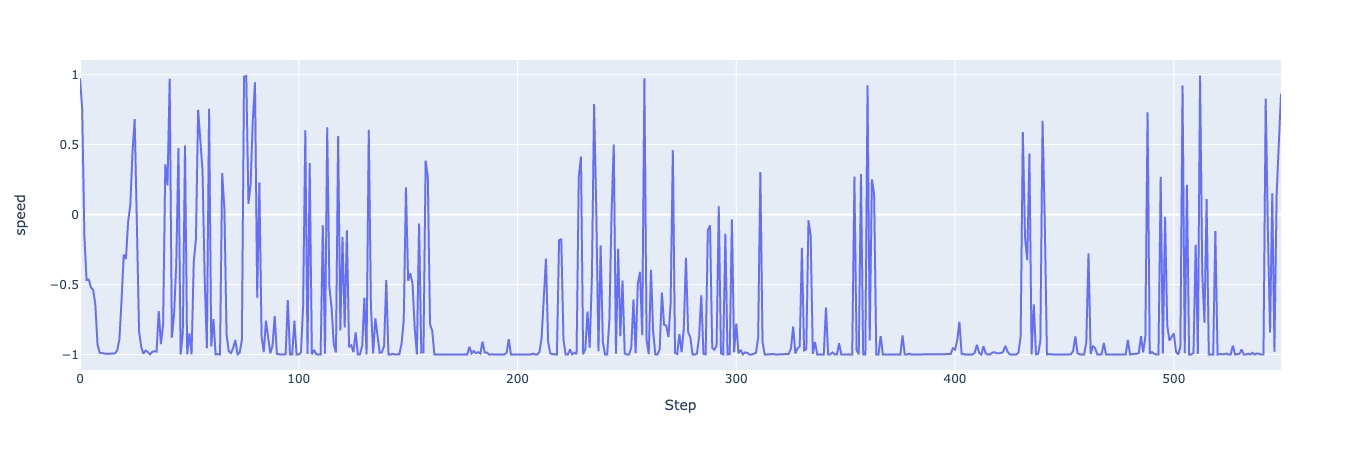}
        \caption{Speed of the car without CAPS.}
    \label{fig:speed_caps}
    \end{subfigure}
    \begin{subfigure}[t]{0.95\columnwidth}
        \includegraphics[width=\columnwidth]{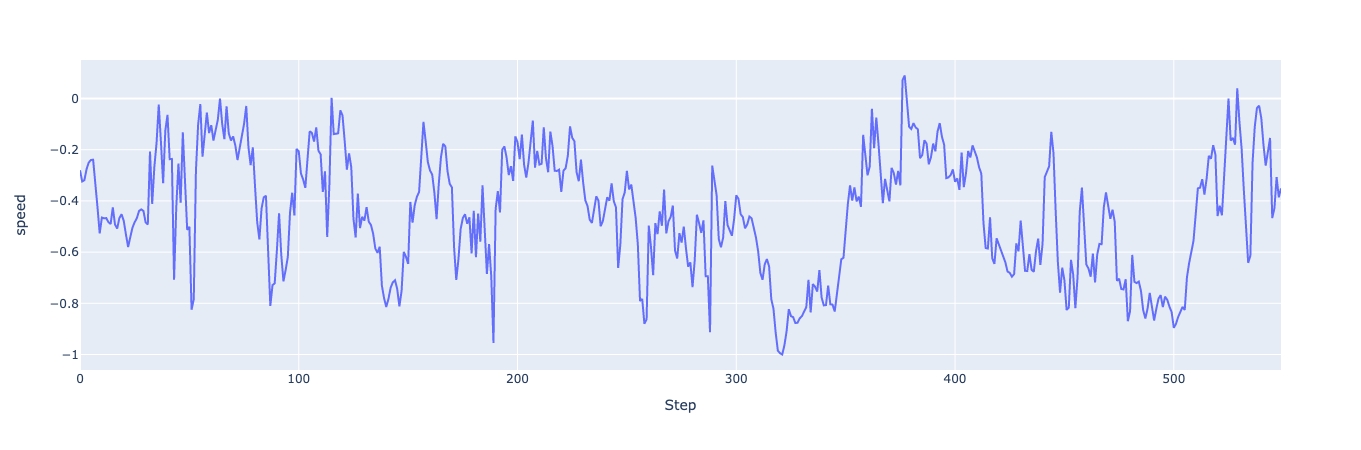}
        \caption{Speed of the car with CAPS.}
        \label{fig:speed_no_caps}
    \end{subfigure}
\caption{Speed of the car without CAPS and with CAPS.}
\label{fig:compare_speed_caps}
\end{figure}

\begin{figure}[t!]
\centering
    \begin{subfigure}[t]{0.95\columnwidth}
        \includegraphics[width=\columnwidth]{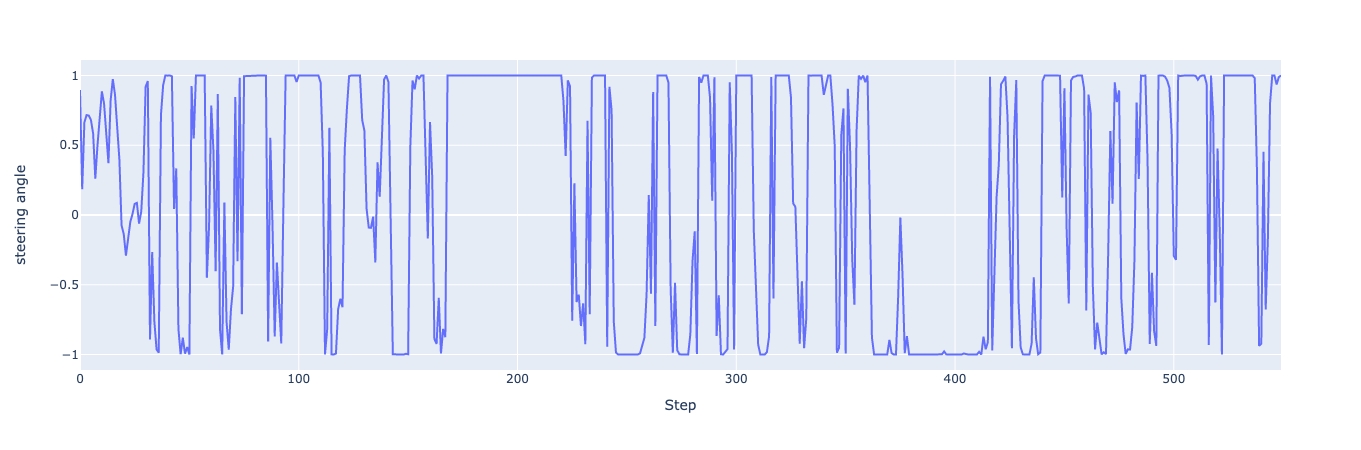}
        \caption{Steering angle of the car without CAPS.}
    \label{fig:steering_caps}
    \end{subfigure}
    \begin{subfigure}[t]{0.95\columnwidth}
        \includegraphics[width=\columnwidth]{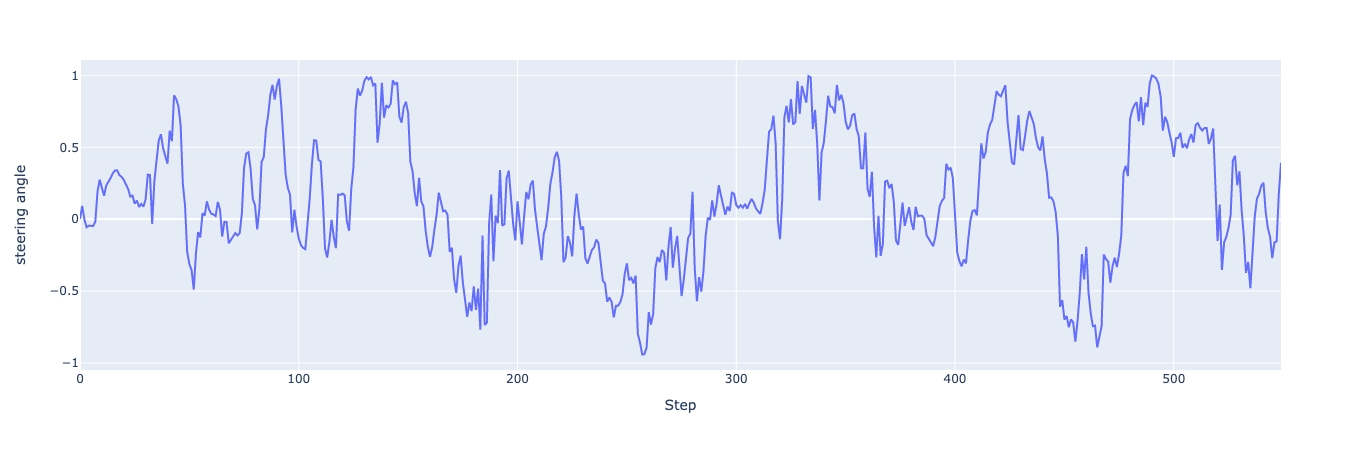}
        \caption{Steering angle of the car with CAPS.}
        \label{fig:steering_no_caps}
    \end{subfigure}
\caption{Steering angle of the car without CAPS and with CAPS.}
\label{fig:compare_steering_caps}
\end{figure}

Figure \labelcref{fig:compare_speed_smooth_caps,fig:compare_steering_smooth_caps,fig:compare_speed_caps,fig:compare_steering_caps} show the comparison of the steering angle and the speed with CAPS and without CAPS.
The model with CAPS achieves more stable behaviors in real-world testing.
\autoref{tab:caps_vs_no_caps} compares the smoothness value of steering angle and speed of the model with CAPS and without CAPS.
The results show that CAPS significantly improves the smoothness value in both speed and steering angle.

%========= TABLES ==============================
%COMPLETION RATE TABLE
\input{tables}
%========= END TABLES ==============================

\subsection{Smoothness Control and Sim-to-Real Policy Transfer}
\label{subsec:caps_cyclegan}
In this experiment, we compare the effectiveness of both the action smoothing method and sim-to-real transfer methods.
We conduct the experiments to compare the following models:
\begin{enumerate}
    \item SAC only (without CAPS, without sim-to-real transfer method).
    \item SAC + CAPS (without sim-to-real transfer method)
    \item SAC + DR (without CAPS)
    \item SAC + DR + CAPS
    \item SAC + CycleGAN (without CAPS)
    \item SAC + CycleGAN + CAPS
\end{enumerate}

\begin{figure}[!htb]
    \centering
        \includegraphics[width=0.95\columnwidth]{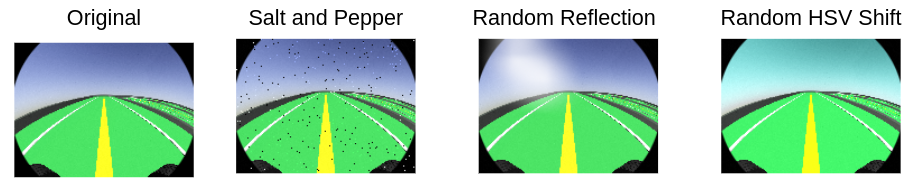}
    \caption{Domain randomization methods for sim-to-real transfer.}
    \label{fig:domain_randomization_result}
\end{figure}

\begin{figure}[!htb]
    \centering
    \begin{subfigure}[t]{0.8\columnwidth}
        \includegraphics[width=\columnwidth]{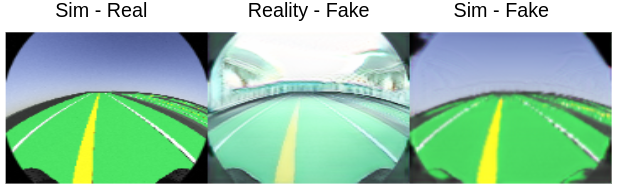}
    \caption{From left to right: the simulation image; the generated real image by CycleGAN; the reconstructed simulation image. }
    \label{fig:sim_real_sim}
    \end{subfigure}
    \begin{subfigure}[t]{0.8\columnwidth}
        \includegraphics[width=\columnwidth]{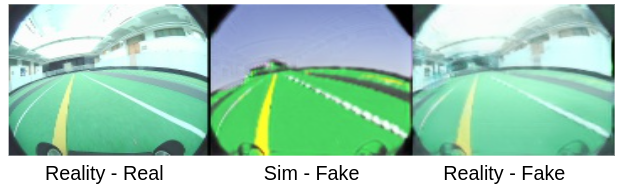}
    \caption{From left to right: the real image; the generated sim image by CycleGAN; the reconstructed real image. }
    \label{fig:real_sim_real}
    \end{subfigure}
    \caption{CycleGAN results.}
    \label{fig:cyclegan_results}
\end{figure}

For the domain randomization methods, we apply three types: Salt and Pepper noise, Random Reflection, Random HSV Shift.
\autoref{fig:domain_randomization_result} shows examples of different domain randomization methods.
For the domain adaptation method, we implement CycleGAN. \autoref{fig:cyclegan_results} shows our CycleGAN result for translating between the simulation and real images and vice versa.
The agent is trained in the simulation with the sim-to-real transfer methods and then tested 15 times with three different speed configurations (denoted as c1, c2, c3) in the real environment. The speed configuration is a range of speed values that the car can take.
We want to test the ability to stable the control of CAPS by increasing the speed of the car.
That is no doubt that running at a higher speed is more difficult for the agent to make a decision.
We setup different speed configurations as follow:
\begin{enumerate}
    \item Real World (c1): from 1.125 to 8 (m/s).
    \item Real World (c2): from 1.125 to 9  m/s.
    \item Real World (c3): from 3.15 to 9.3 (m/s)
\end{enumerate}

\autoref{tab:completion_rate} and \autoref{tab:finishing_lap_time} show the results of completion rate and average finish lap time.
Without CAPS and without sim-to-real transfer method, the agents can not transfer the policies from the simulation to the real environment.
With CAPS, the  behaviors of the agents are more stable, therefore can finish a run faster than the agents without CAPS.


\subsection{Study the impact of CAPS components}
\label{subsec:impact_CAPS}
In this experiment, we want to study the impact of each component on the objective function of CAPS (\autoref{equation:objective_functino_caps}).
We compare the finishing lap time and the completion rate of the following models:
\begin{enumerate}
    \item SAC only.
    \item SAC + Steering angle penalty: a reward engineering method with Steering Angle Penalty = 0.003 * abs(degree).
    \item SAC + Temporal Smoothness term. 
    \item SAC + Spatial Smoothness term.
    \item SAC + CAPS (Spatial + Temporal).
\end{enumerate}

\begin{figure}[t!]
\centering
    \begin{subfigure}[t]{0.8\columnwidth}
        \includegraphics[width=\columnwidth]{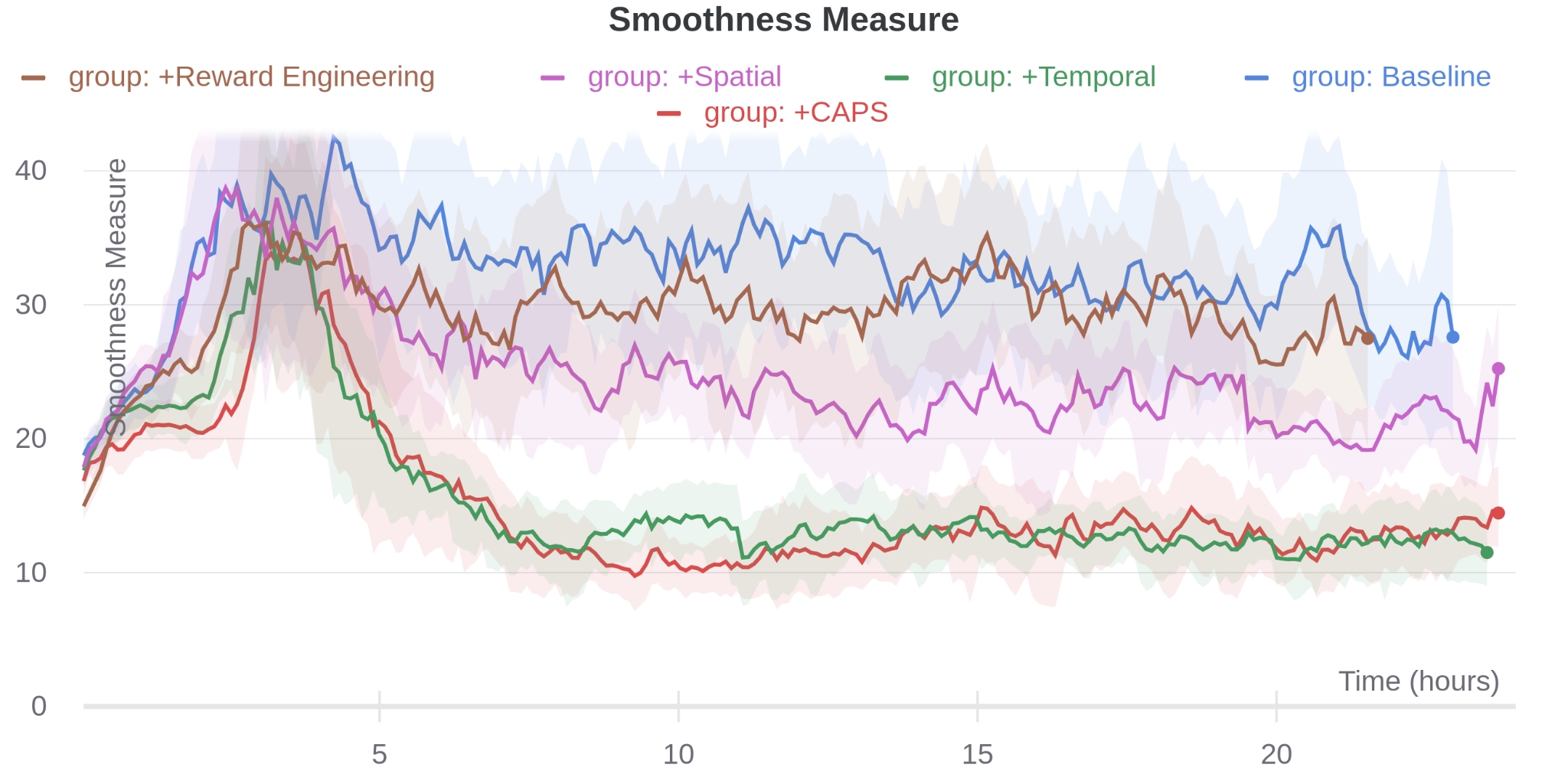}
        \caption{Smoothness value.}
    \label{fig:caps_completion}
    \end{subfigure}
    \begin{subfigure}[t]{0.8\columnwidth}
        \includegraphics[width=\columnwidth]{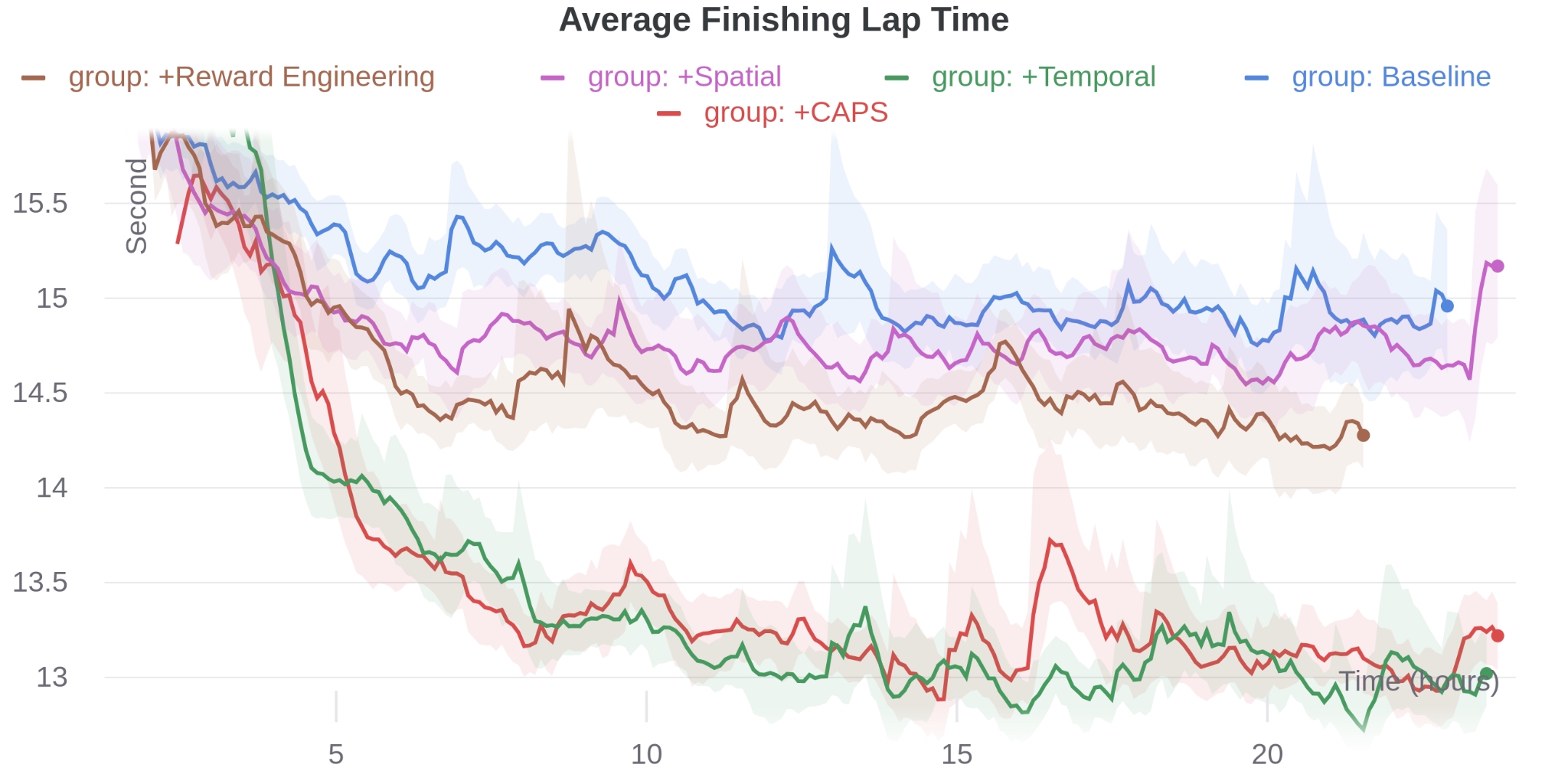}
        \caption{Average finishing lap time.}
        \label{fig:caps_finish}
    \end{subfigure}
\caption{Compare the impact of the components in CAPS.}
\label{fig:compare_caps_component}
\end{figure}

\autoref{fig:compare_caps_component} shows the result of comparison. We can see that the Temporal Smoothness or the Spatial Smoothness individually improves the smoothness value.
Moreover, the Temporal Smoothness takes the most impact to the result and then significantly reduces the finishing lap time.

\subsection{Temporal Smoothness Sensitivity Analysis}
\label{subsec:Temporal_Smoothness_Sensitivity}
From the experiment above (\ref{subsec:impact_CAPS}), we see that the Temporal Smoothness has the most impact on improving the agent's performance.
In this experiment, we analyze the sensitivity of the Temporal Smoothness in CAPS.
We remove the Spatial Smoothness term by setting the $\lambda_{S} = 0$ and then test with different values of $\lambda_{T}$: 0.5, 0.8, 1.0, 1.3.

\begin{figure}[t!]
\centering
    \begin{subfigure}[t]{0.8\columnwidth}
        \includegraphics[width=\columnwidth]{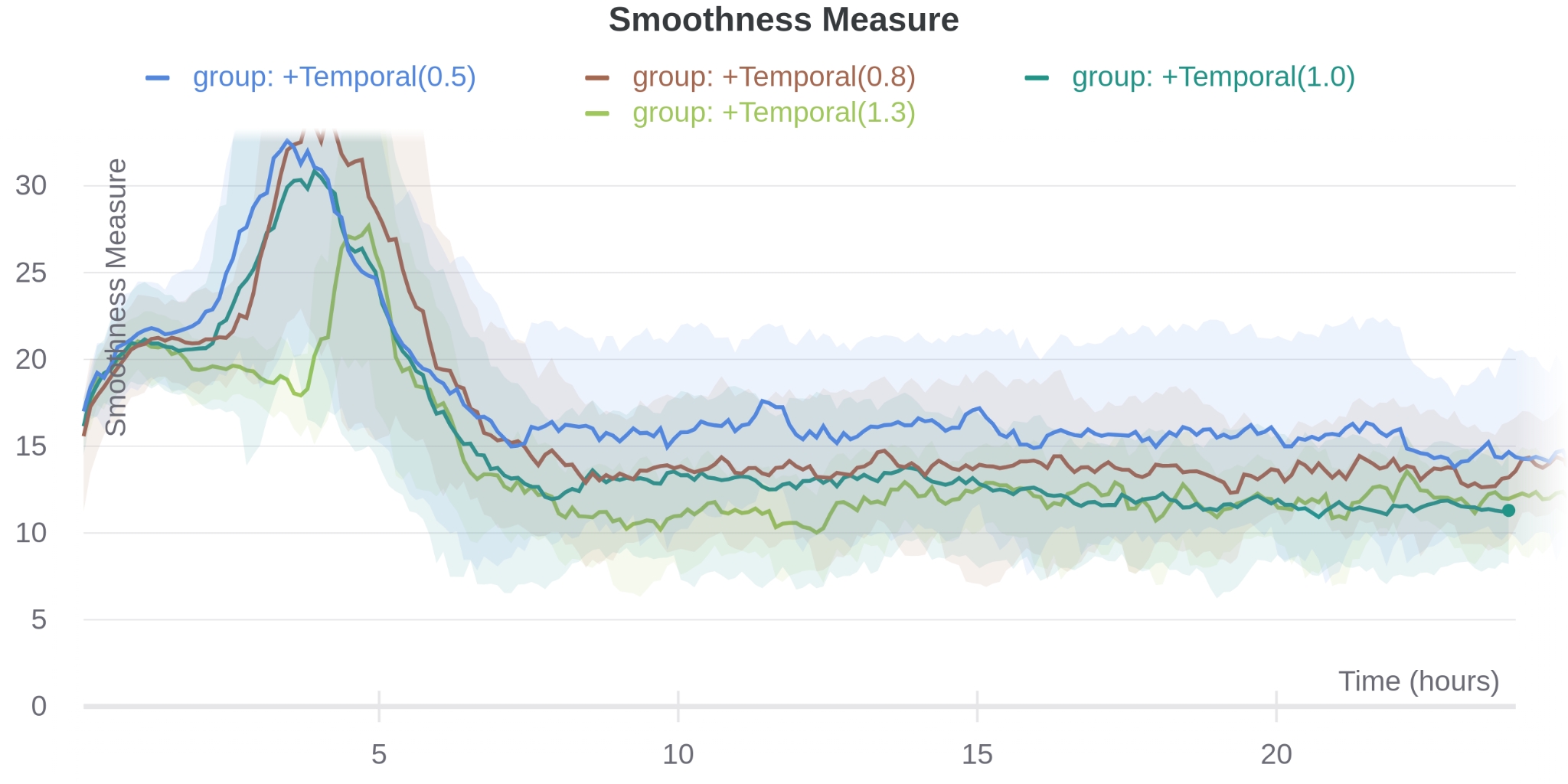}
        \caption{Smoothness value.}
    \label{fig:caps_temporal_sensitive_smoothness}
    \end{subfigure}
    \begin{subfigure}[t]{0.8\columnwidth}
        \includegraphics[width=\columnwidth]{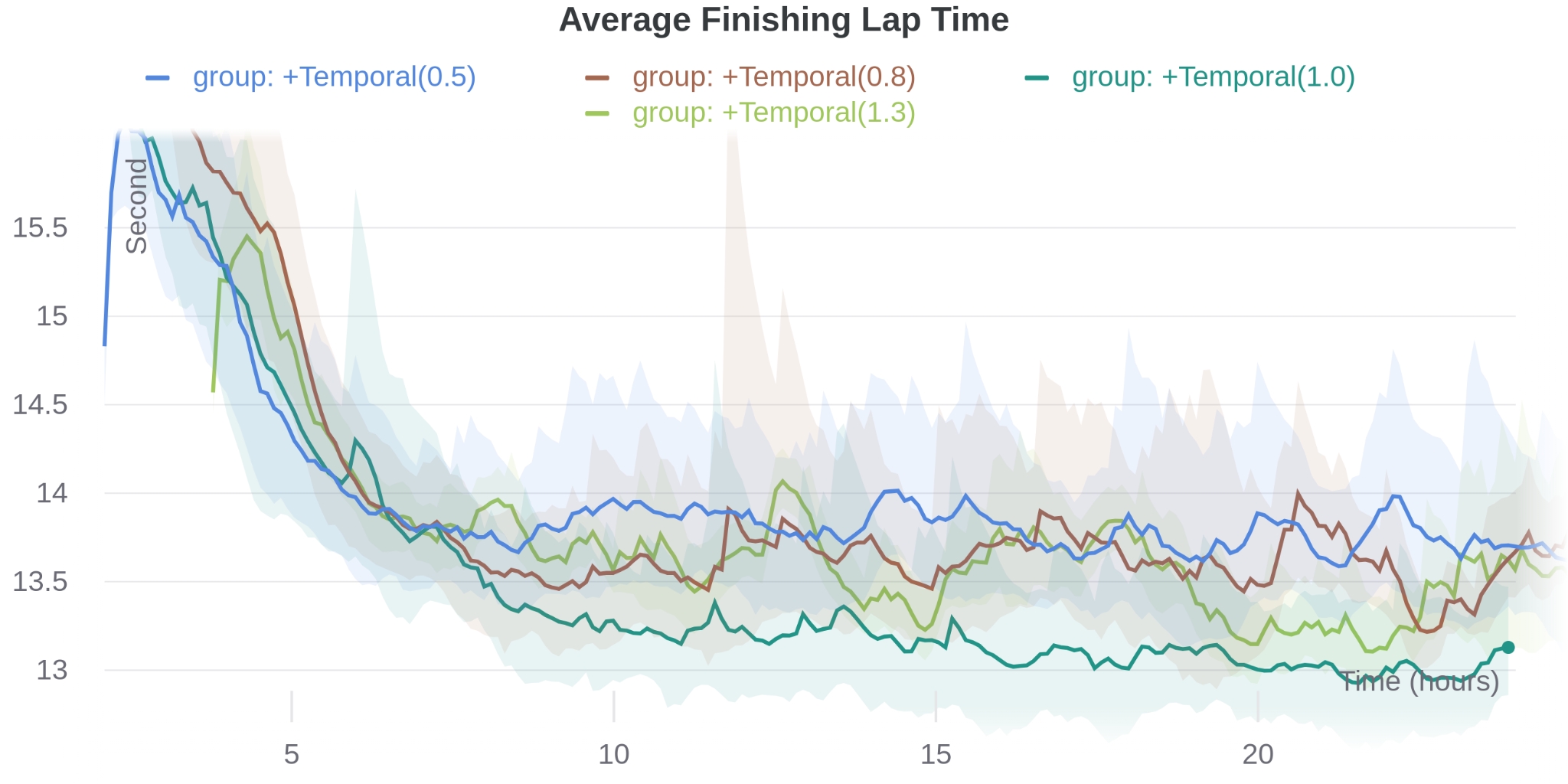}
        \caption{Average finishing lap time.}
        \label{fig:caps_temporal_sensitive_finish}
    \end{subfigure}
\caption{Compare the sensitivity of Temporal Smoothness.}
\label{fig:compare_temporal_sensitiviy}
\end{figure}

From the results in \autoref{fig:compare_temporal_sensitiviy}, we see that $\lambda_{T}=1.0$ gives the best result for both smoothness value and finishing lap time. 

\subsection{Spatial Smoothness Ablation Study}
\label{subsec:ablation_study}
\begin{figure}[t!]
\centering
    \begin{subfigure}[t]{0.8\columnwidth}
        \includegraphics[width=\columnwidth]{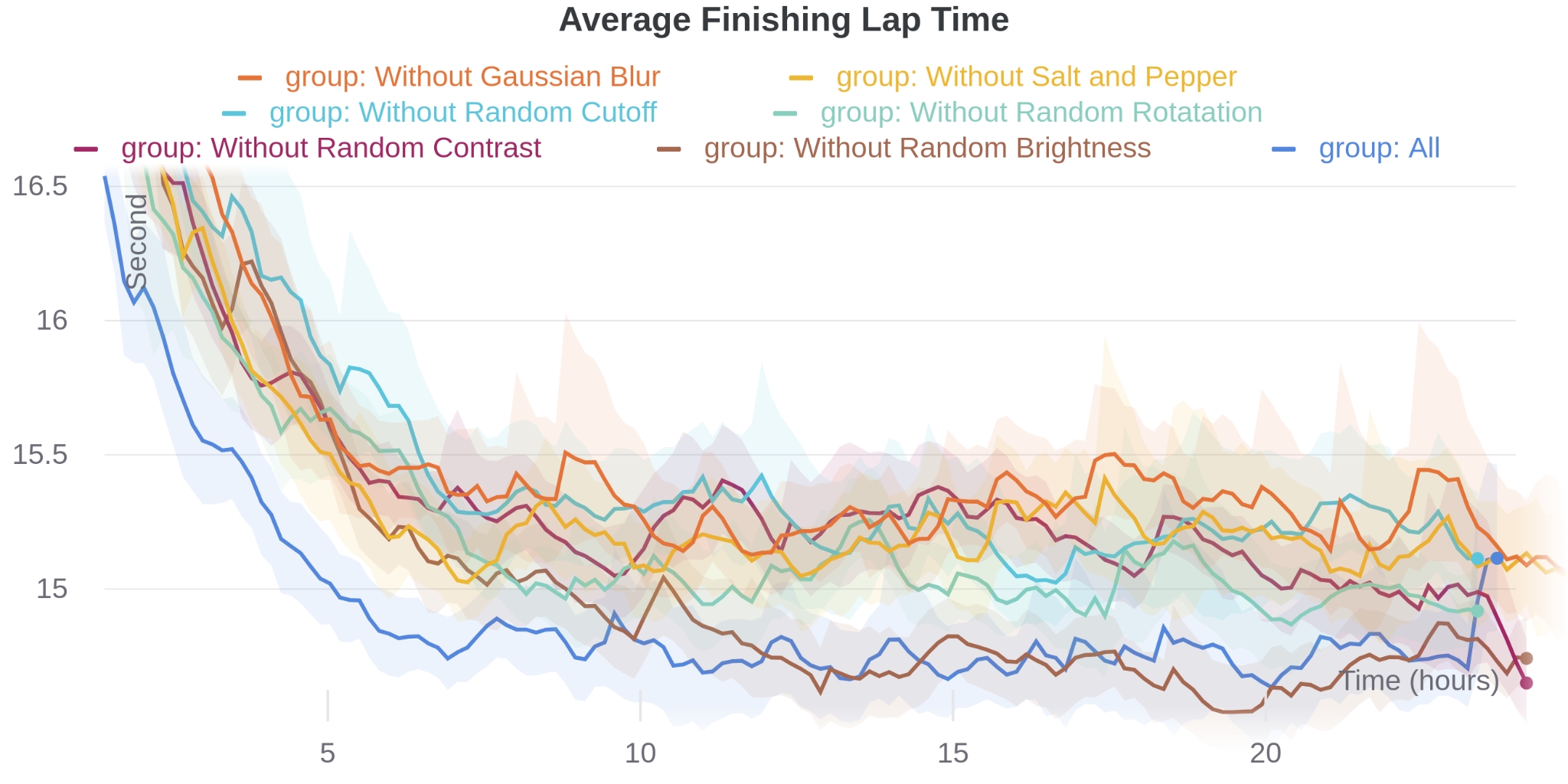}
        \caption{Finishing lap time in Spatial Smoothness ablation study.}
    \label{fig:ablation_finish}
    \end{subfigure}
    \begin{subfigure}[t]{0.8\columnwidth}
        \includegraphics[width=\columnwidth]{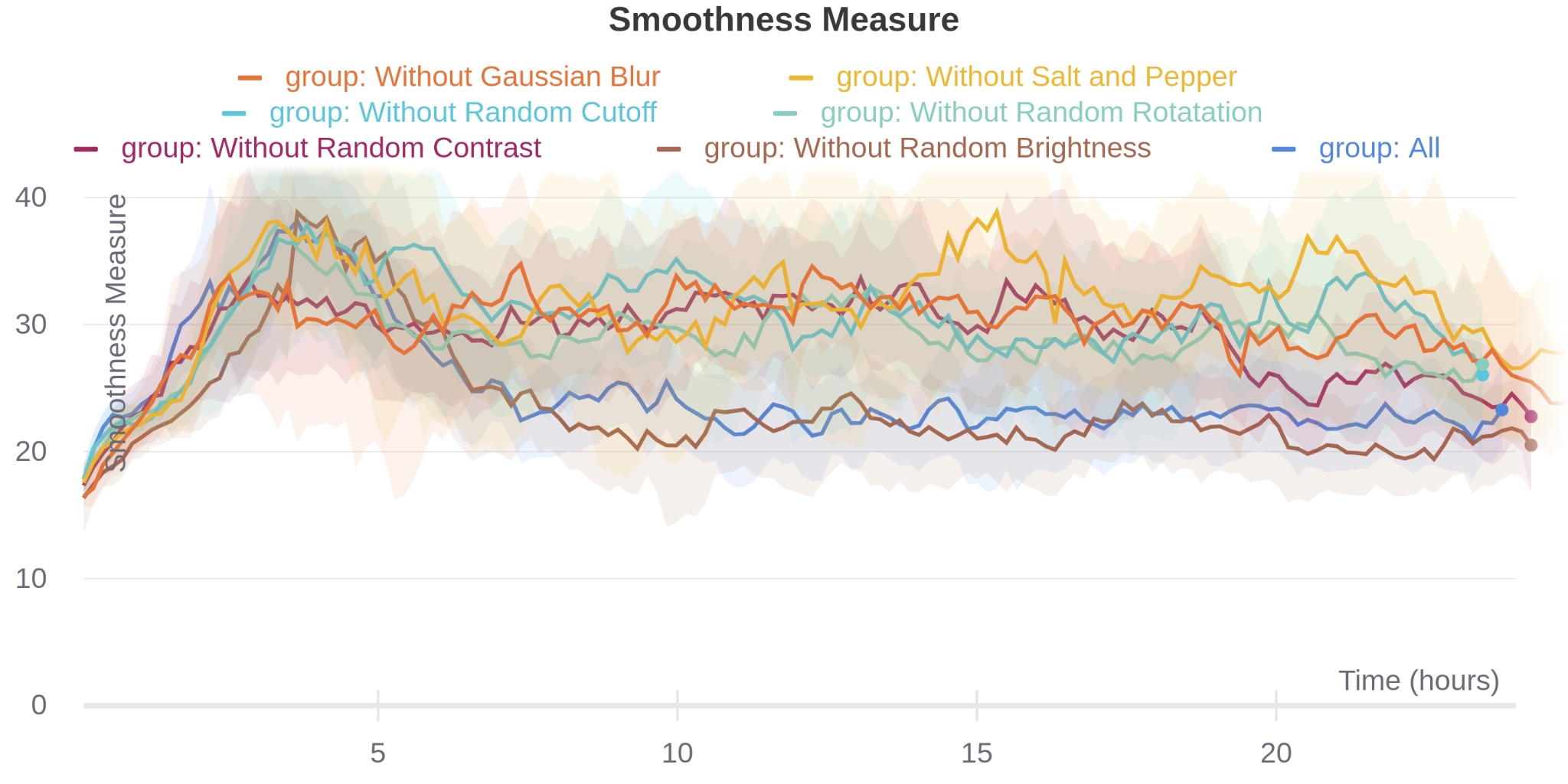}
        \caption{Finishing lap time in Spatial Smoothness ablation study.}
        \label{fig:ablation_smoothness}
    \end{subfigure}
\caption{Spatial Smoothness ablation study.}
\label{fig:ablation}
\end{figure}

We conduct an ablation study to investigate the influence of the randomization methods on Spatial Smoothness.
In this experiment, we individually remove each randomization method to compare the impact of each method on the Spatial Smoothness.
The experiment shows that the performance is less reduced when removing the Random Brightness while removing the Gaussian Blur or Salt and Pepper clearly drop the result.
A possible explanation is that the image captured in the simulation has less noise than the image captured in the real world.
Adding Salt and Pepper helps the agent cover this type of image.
Moreover, when the car runs fast, the captured images are easily blurred; thus, training with more blurred images improves the performance of the agent.
\autoref{fig:ablation} show the result of the Spatial Smoothness ablation study.

%===============================================================================
% \section{Conclusion}
% This paper presents image-based CAPS, an image-based regularizing action policies method to smooth the control of autonomous miniature car racing.
% We then combine image-based CAPS with sim-to-real methods to transfer the trained policies from the simulation to the real world.
% The model that combines image-based CAPS and CycleGAN achieved the best result in real-world testing, which reduces 21.80\% of the average finishing lap time.
% Comparing the effectiveness of different sim-to-real transfer methods combined with CAPS, we find that the CycleGAN has better completion rates than the other domain randomization methods.
% % Through our extensive experiments, we analyze the influence of each component in CAPS.
% Analyzing the impact of each component in CAPS show that the Temporal Smoothness term takes the most impact on the performance of CAPS; and the Gaussian Blur is the most important randomization method to train the Spatial Smoothness term.

\section{Conclusion}
This paper presents image-based CAPS, an image-based regularizing action policies method to smooth the control of autonomous miniature car racing.
% We then combine image-based CAPS with sim-to-real methods to transfer the trained policies from the simulation to the real world.
The model that combines image-based CAPS and CycleGAN achieved the best result in real-world testing, which reduces the average finishing lap time; while improving the completion rate.
% Comparing the effectiveness of different sim-to-real transfer methods combined with CAPS, we find that the CycleGAN has better completion rates than the other domain randomization methods.
% % Through our extensive experiments, we analyze the influence of each component in CAPS.
Analyzing the CAPS components shows that the Temporal Smoothness term takes more impact on the performance of CAPS.
An ablation study of the Spatial Smoothness term indicates that Salt and Pepper and Gaussian Blur are two important randomization methods influencing the result.

%===============================================================================

%%%%%%%%APPENDIX%%%%%%%%%%%%%%%%%%%%%%%%%%%%%%%%%%%%%%%%%%%%%%%%%%%%%%%%%%%%%%%%%%%%%%%%


%%%%REFERENCES%%%%%%%%%%%%%%%%%%%%%%%%%%%%%%%%%%%%%%%%%%%%%%%%%%%%%%%%%%%%%%%%%%%%%%%%%%%%

\bibliography{example.bib}  % .bib
\end{document}

%% file: tables.tex
%========= TABLES ==============================
%COMPLETION RATE TABLE

\begin{table*}[!t]
\begin{tabular}{|l|cc|cc|cc|}
\hline
\multicolumn{1}{|r|}{} & \multicolumn{2}{c|}{Baseline}             & \multicolumn{2}{c|}{Domain Randomization} & \multicolumn{2}{c|}{CycleGAN}                    \\ \hline
                       & \multicolumn{1}{c|}{w/o CAPS} & with CAPS & \multicolumn{1}{c|}{w/o CAPS} & with CAPS & \multicolumn{1}{c|}{w/o CAPS} & with CAPS        \\ \hline
Real World (c1) & \multicolumn{1}{c|}{0\%} & 0\% & \multicolumn{1}{c|}{0\%} & \textbf{100\%} & \multicolumn{1}{c|}{73.33\%}          & 73.33\% \\ \hline
Real World (c2) & \multicolumn{1}{c|}{0\%} & 0\% & \multicolumn{1}{c|}{0\%} & 26.67\%        & \multicolumn{1}{c|}{\textbf{86.67\%}} & 80.00\% \\ \hline
Real World (c3)        & \multicolumn{1}{c|}{0\%}      & 0\%       & \multicolumn{1}{c|}{0\%}      & 0\%       & \multicolumn{1}{c|}{13.33\%}  & \textbf{20.00\%} \\ \hline
\end{tabular}
\caption{Completion rate.}
\label{tab:completion_rate}
\end{table*}
%FINISH LAP TIME TABLE
\begin{table*}[!t]
\begin{tabular}{|l|cc|cc|cc|}
\hline
\multicolumn{1}{|r|}{} & \multicolumn{2}{c|}{Baseline}  & \multicolumn{2}{c|}{Domain Randomization} & \multicolumn{2}{c|}{CycleGAN}                 \\ \hline
 &
  \multicolumn{1}{l|}{w/o CAPS} &
  \multicolumn{1}{l|}{with CAPS} &
  \multicolumn{1}{l|}{w/o CAPS} &
  \multicolumn{1}{l|}{with CAPS} &
  \multicolumn{1}{l|}{w/o CAPS} &
  \multicolumn{1}{l|}{with CAPS} \\ \hline
Real World (c1)        & \multicolumn{1}{c|}{NaN} & NaN & \multicolumn{1}{c|}{NaN}     & 32.47s     & \multicolumn{1}{c|}{22.20s} & \textbf{17.36s} \\ \hline
Real World (c2)        & \multicolumn{1}{c|}{NaN} & NaN & \multicolumn{1}{c|}{NaN}     & 24.33s     & \multicolumn{1}{c|}{19.77s} & \textbf{16.05s} \\ \hline
Real World (c3)        & \multicolumn{1}{c|}{NaN} & NaN & \multicolumn{1}{c|}{NaN}     & NaN        & \multicolumn{1}{c|}{17.20s} & \textbf{14.71s} \\ \hline
\end{tabular}
\caption{Average finishing lap time.}
\label{tab:finishing_lap_time}
\end{table*}

%========= END TABLES ==============================